\documentclass[letterpaper,10pt,conference]{ieeeconf}
\IEEEoverridecommandlockouts       
\usepackage{amssymb}
\usepackage{amsmath}
\usepackage[english]{babel}
\usepackage{float}
\usepackage{graphicx}
\usepackage{hyperref}
\usepackage{mathtools}

\usepackage[noadjust]{cite}
\usepackage[font=footnotesize,labelfont=footnotesize]{caption}
\usepackage{comment}
\usepackage{authblk}
\usepackage{multirow}
\usepackage{xcolor}
\usepackage{algorithm}
\usepackage{algorithmic}
\usepackage{xspace}
\usepackage{balance}
\usepackage{booktabs}
\usepackage{array}
\usepackage{dsfont}

\definecolor{darkblue}{rgb}{0.15,0.15,0.55}
\definecolor{lightgrey}{rgb}{0.75,0.75,0.75}

\newcommand\revision[1]{{\color{black}{#1}}}

\bstctlcite{RAL:BSTcontrol}

\hypersetup{
  pdftitle={Dec-MARVEL: Decentralized Multi-Agent Exploration without Communication under Budget Constraints},
  pdfauthor={Janghyun Cho; Jimmy Chiun; Guillaume Sartoretti; Changjoo Nam}
}

\begin{document}
\title{\LARGE \bf  Dec-MARVEL: Decentralized Multi-Agent Exploration without Communication under Budget Constraints\vspace{-7pt}}
\author{Janghyun Cho$^{1,*}$, Jimmy Chiun$^{2,*}$, Guillaume Sartoretti$^{2}$, and Changjoo Nam$^{3,\dagger}$%
\thanks{$^{*}$Equal contribution.}%
\thanks{$^{\dagger}$Corresponding author: {\tt\small cjnam@sogang.ac.kr}}%
\thanks{$^{1}$Dept. of Artificial Intelligence, Sogang University.}%
\thanks{$^{2}$Dept. of Mechanical Engineering, College of Design and Engineering, National University of Singapore.}%
\thanks{$^{3}$Dept. of Electronic Engineering, Sogang University.}%
}
% \thanks{This work was supported in part by Samsung Research Funding \& Incubation Center of Samsung Electronics under Project Number SRFC-TD2103-01 and the National Research Foundation of Korea (NRF) grant funded by the Korea government (MSIT) (No. 2022R1C1C1008476).
% $^1$Dept. of Electronic Engineering, Sogang University. $^*$Corresponding author: {\tt\small cjnam@sogang.ac.kr}}
% }

%This work was supported by the Sogang University Research Grant of 2022.
\maketitle

\begin{abstract}
Multi-UAV exploration is often constrained by unreliable communication, limited field-of-view sensing (e.g., lightweight onboard camera), 
%\janghyun{and finite travel budgets that must be reserved enough to return to its base.}
and finite travel budgets that require each robot to reserve enough budget to return to its base.
\revision{We present Dec-MARVEL, a decentralized, communication-free exploration policy that jointly selects waypoints and sensing headings from each robot’s local map, remaining budget, and opportunistic observations of teammates.}
% We present Dec-MARVEL, a decentralized budget-aware exploration framework for communication-free teams with directional sensing. 
% Rather than exchanging maps, goals, or messages, each robot coordinates through its incidental observations: any teammate trajectory within its field of view serves as a coordination signal.
% \janghyun{Each robot coordinates without exchanging maps, goals, or messages. Instead, it relies only on teammate trajectories that happen to fall within its field of view as an implicit coordination signal.}
% \janghyun{A graph-attention actor maps local frontier geometry, observed teammate motion, and budget state to return-feasible waypoint-heading actions. We train it with a mixture-based budget curriculum and phase-conditioned critics: one is a training-only privileged critic that exploits the merged team belief.}
% A graph-attention actor fuses local frontier geometry, teammate motion, and budget features to select return-feasible waypoint-heading actions. The actor is trained with phase-conditioned critics, a training-only task-oriented privileged critic, and a mixture-based budget curriculum.
\revision{Across 900 held-out trials spanning three team sizes (2, 4, 8 robots) and three travel budgets (720, 800, and 1024\,m) against four baselines, \mbox{Dec-MARVEL} achieves the highest or tied-highest exploration and success rates and the shortest reported travel to 99\% merged coverage among successful trials in all nine budget--team configurations. Under our tightest 720\,m budget, it reaches 46\%, 84\%, and 94\% success for 2, 4, and 8 robots, versus 17\%, 57\%, and 85\% for the strongest baseline in each case.} Physical-robot experiments show that the resulting waypoint-heading plans are executable and remain budget-feasible.
\end{abstract}

\section{Introduction}
\vspace{-3pt}

Autonomous exploration of unknown environments is a fundamental problem in field robotics, with applications in search and rescue, infrastructure inspection, and hazardous-environment surveys. Using multiple robots can improve exploration efficiency by covering different regions in parallel~\cite{yamauchi1998frontier}. However, this benefit depends strongly on coordination: \revision{if multiple robots explore independently, they tend to cover the same regions redundantly, which limits the benefit of parallelism.} Many existing multi-robot exploration methods address this issue by relying on explicit communication, such as sharing maps, exchanging goals, or assigning frontiers among robots~\cite{yu2021smmr,zhou2023racer}. In practical fields, especially on uncrewed aerial vehicles (UAVs), this assumption can be difficult to satisfy because wireless communication can be degraded or unavailable due to occlusions, signal attenuation, limited bandwidth, or infrastructure damage~\cite{da2024communication,tan2024ir}.

\begin{figure}[t]
\vspace{-5pt}
\captionsetup{skip=2pt}
\centering
\includegraphics[width=0.99\columnwidth]{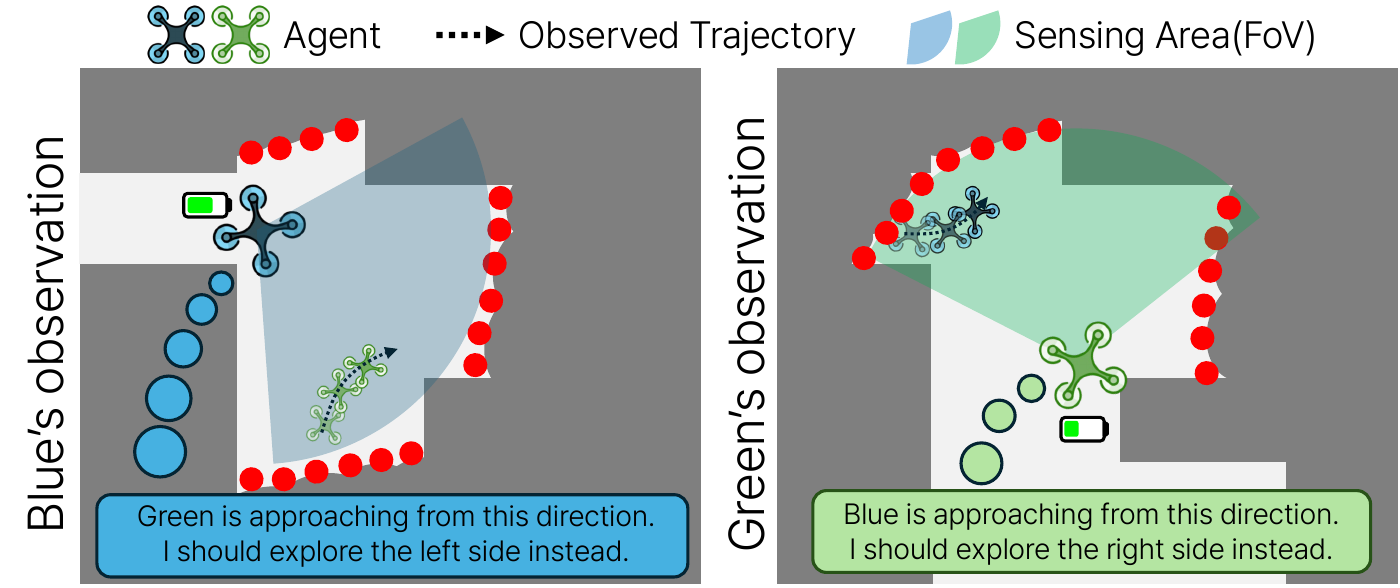}
\caption{Dec-MARVEL uses locally observed teammate trajectories to guide budget-aware waypoint-heading decisions without inter-robot messages.}
\label{fig:representative}
\vspace{-26pt}
\end{figure}

In this work, we consider a different form of coordination. Even when robots do not exchange messages, they may still occasionally observe one another through their onboard sensors. Prior work on vision-based and limited Field-of-View (FoV) coordination suggests that such visibility constraints can be used as part of the coordination process~\cite{schilling2019learning,catellani2023distributed,ghanta2025space}. These opportunistic observations (Fig.~\ref{fig:representative}) are limited and intermittent.
% , but we show that they can still provide useful information about where a teammate has recently moved, and thus implicitly about which regions may have already been explored. 
This motivates the main question of this paper: \textit{Can visually observed teammate trajectories be used as a coordination signal for multi-robot exploration without requiring map sharing, goal exchange, or explicit task allocation?}

This question is further complicated by two deployment constraints that are common in multi-UAV systems. First, many UAVs use directional sensors, such as forward-facing cameras or limited-angle LiDAR, meaning that robots must decide not only where to move but also where to orient their sensor (where to look)~\cite{chiun2025marvel,catellani2023distributed}. Second, each robot has a finite travel budget (often, tied to battery life).
% and must preserve enough remaining budget to return to base and upload the gathered information during the mission. 
% 
%\janghyun{Communication-free exploration in this setting therefore requires joint reasoning over motion, sensing direction, observed teammate behavior, and remaining travel budget.}
Together, these constraints make communication-free exploration a problem of joint reasoning over motion, sensing direction, observed teammate behavior, and remaining travel budget.

% Decentralized Multi-Agent Exploration without Communication under Budget Constraints
Aimed at those challenges, we propose Dec-MARVEL.
%, a communication-free, decentralized budget-aware exploration framework that uses opportunistically observed teammate trajectories as the only deployment-time coordination cue.
\revision{Each robot selects its next waypoint and sensing heading from its local frontiers, remaining travel budget, and visually observed teammate trajectories, without communication, while a budget guard ensures that it can return before exhausting its budget.} 
% During training, a centralized learning strategy and a budget curriculum improve coordination and robustness from privileged team information, while the deployed policy remains fully decentralized and communication-free. 
\revision{We show in simulation that this coordination signal improves success and coverage efficiency under tight travel budgets, and demonstrate on physical robots that the resulting plans are executable within the travel budget.}
The contributions of this work are summarized as follows:

\begin{itemize}
    \item \textbf{Communication-free, budget-aware exploration:}
    We introduce Dec-MARVEL, which jointly selects waypoints and sensing headings from each robot's local map, remaining budget, and opportunistically observed teammate trajectories, while enforcing return feasibility without exchanging maps, goals, actions, or messages.

    \item \textbf{Trajectory-conditioned policy learning:}
    We combine per-timestep teammate-trajectory encoding with a task-oriented privileged (TOP) critic, phase-conditioned critics, and a mixture-based budget curriculum for decentralized deployment across different travel budgets.

    \item \textbf{Extensive validation:}
    Across 900 simulation trials, Dec-MARVEL achieves the highest or tied-highest exploration and success rates and the shortest reported travel to 99\% coverage among successful trials in all nine budget--team settings. Dropout and perturbation tests evaluate sparse/noisy teammate observations, while physical-robot experiments validate budget-feasible plan execution.
\end{itemize}

\section{Related Work}

\subsection{Multi-Robot Exploration}
\vspace{-2pt}
\revision{Frontier-based methods~\cite{yamauchi1997frontier, yamauchi1998frontier, keidar2014efficient} select free-unknown boundaries as targets. Sampling-based planners optimize information gain over viewpoints~\cite{bircher2016receding, duberg2022ufoexplorer}.
Recent learning-based methods improve adaptivity, e.g., decentralized graph-attention policies under FoV constraints~\cite{chiun2025marvel} and macro-action policies under communication disruptions~\cite{tan2022deep}.} % reduced additional illustration and citation
\revision{Most of these formulations treat exploration as fixed-horizon coverage, whereas Dec-MARVEL jointly selects where to move and where to sense under a hard return-budget constraint.}

\subsection{Communication-Constrained Coordination}
\vspace{-3pt}
% \revision{Coordination is essential in multi-robot exploration because independent local decisions can lead to redundant coverage and poor scaling with team size.}
Many systems reduce communication load rather than remove communication entirely: SMMR-Explore~\cite{yu2021smmr} exchanges submaps, RACER~\cite{zhou2023racer} uses decentralized task allocation, and RAMEN~\cite{zhao2025ramen} and UDON~\cite{zhao2025udon} address robust neural implicit mapping under degraded links. 
% \revision{These methods make multi-robot exploration more practical under bandwidth or reliability limits, but they still assume some mechanism for sharing information, allocating tasks, or synchronizing beliefs.}
Other methods handle intermittent communication through rendezvous or reconnection strategies~\cite{da2024communication,tan2024ir}, while communication-denied settings have considered frontier allocation from previously shared observations~\cite{bautin2011towards}, minimal signaling in UAV swarms~\cite{mcguire2019minimal}, or LiDAR-based communication-free collective navigation~\cite{choi2026communication}. 
% revision{In contrast, Dec-MARVEL assumes no inter-robot communication during deployment; coordination is induced only through teammate trajectories opportunistically observed within each robot's vision.}

\subsection{Implicit Coordination and Directional Sensing}
\vspace{-3pt}
Several methods augment local beliefs using predicted or inferred information, including diffusion-based map prediction~\cite{tan20264cnet}, uncertainty-aware viewpoint selection~\cite{ho2025mapex}, epistemic planning for restricted communication~\cite{bramblett2023epistemic}, 
\revision{and state modeling for communication-free cooperative MARL~\cite{kontogiannis2025enhancing}.}
%revision{These approaches show that planning can benefit from information that is not directly observed, either through prediction, uncertainty estimation, or belief reasoning.}
Vision-based coordination has also been studied for UAV swarms~\cite{schilling2019learning}, and limited-FoV control has been addressed~\cite{catellani2023distributed}. \revision{None of these works combines opportunistic teammate observation, coupled waypoint-heading selection, and hard return-budget constraints in a single communication-free policy. In this setting, the directional sensor determines both which frontiers can be explored and which teammates can be observed, and Dec-MARVEL learns to exploit the resulting weak cue for coordinated, budget-aware sensing decisions.}

% \subsection{Trajectory Representation}
% \vspace{-3pt}
% Transformer-based motion models encode temporal agent histories with attention and positional encodings~\cite{yuan2021agentformer,shi2022motion,zhang2023trafficbots}, and gated attention can suppress low-information  heads~\cite{qiu2026gated}. Dec-MARVEL adapts these prediction-oriented encoders to short, intermittent teammate trajectories for exploration control.

\subsection{Privileged Training and Budget Curriculum}
\vspace{-3pt}
Privileged training is standard in robot learning, from asymmetric actor-critics trained on full simulator state ~\cite{pinto2018asymmetric} to centralized critics in cooperative MARL ~\cite{yu2022surprising}, while automatic curricula adapt task difficulty from learning progress ~\cite{portelas2020teacher} and revisit degrading tasks to avoid forgetting ~\cite{matiisen2019teacher}. Dec-MARVEL combines both (Secs.~\ref{sec:top_critic} and~\ref{sec:curriculum}).

\section{Problem Formulation}
\label{sec:prob}
\vspace{-3pt}

\subsection{Budget-Constrained, Communication-Free Dec-POMDP}
\vspace{-3pt}

We formulate exploration as a decentralized partially observable Markov decision process (Dec-POMDP) with hard travel-budget constraints. The team consists of $N$ robots operating in an unknown environment. At time $t$, robot $i$ has pose $p_{i,t}=(x_{i,t},\psi_{i,t})$, local observation $o_{i,t}$, local belief map $M_{i,t}$, and remaining travel budget $B_{i,t}$. The deployed policy is decentralized:
\begin{equation}
\label{eq:policy}
a_{i,t}\sim \pi_\theta(a_{i,t}\mid o_{i,t},B_{i,t}),
\end{equation}
and no actor receives teammate maps, goals, actions, or messages during execution. Deployment-time teammate information is obtained solely through detections satisfying sensing-range, FoV, and line-of-sight constraints, and is encoded as trajectory tokens and teammate-visit evidence.

Motion consumes budget according to traveled distance, and the budget is updated as $ B_{i,t+1}=B_{i,t}-\|x_{i,t+1}-x_{i,t}\|_2 $. 

Here, $x_{i,t}\in\mathbb{R}^2$ is the planar position and $\psi_{i,t}$ is the sensing heading. We model budget as traveled path length; yaw motion and sensing energy are not charged separately.

Let $D_i(x,x_i^0;\mathcal{G}_{i,t})$ denote the shortest-path distance from position $x$ to the base $x_i^0$ on robot $i$'s current graph $\mathcal{G}_{i,t}$. A candidate move from $x_{i,t}$ to $x'$ is return-feasible only if
\begin{equation}
\label{eq:budget_feasibility}
\|x'-x_{i,t}\|_2
+
D_i(x',x_i^0;\mathcal{G}_{i,t})
+\epsilon
\le B_{i,t},
\end{equation}
where $\epsilon$ is a safety margin. If no feasible exploration action exists, or if
\begin{equation}
\label{eq:return_mode_change}
D_i(x_{i,t},x_i^0;\mathcal{G}_{i,t})+\epsilon \geq B_{i,t},
\end{equation}
robot $i$ switches to return mode and follows the shortest graph path back to its base. This constraint is enforced during execution by action filtering and return-mode switching, rather than only through reward penalties.

% \subsection{Budget-Constrained Multi-Agent POMDP}
% \vspace{-3pt}

% We formulate exploration as a decentralized partially observable Markov decision process (Dec-POMDP) with hard travel-budget constraints. Each robot $i$ maintains a pose $p_{i,t}=(x_{i,t},\psi_{i,t})$, a local observation $o_{i,t}$, and a remaining travel budget $B_{i,t}$. Motion consumes budget according to traveled distance:
% \begin{equation}
% \label{eq:budget_update}
% B_{i,t+1}=B_{i,t}-\|x_{i,t+1}-x_{i,t}\|_2.
% \end{equation}

% Here, $x_{i,t}\in\mathbb{R}^2$ is the planar robot position and $\psi_{i,t}$ is the sensing heading. We model budget as traveled path length; yaw motion and sensing energy are not charged separately.

% Let $D_i(x,x_i^0;\mathcal{G}_{i,t})$ denote the shortest-path distance from position $x$ to the base on robot $i$'s current graph $\mathcal{G}_{i,t}$. A candidate move from $x_{i,t}$ to $x'$ is feasible only if
% \begin{equation}
% \label{eq:budget_feasibility}
% \|x'-x_{i,t}\|_2 + D_i(x',x_i^0;\mathcal{G}_{i,t}) + \epsilon \le B_{i,t},
% \end{equation}

% where $\epsilon$ is a configurable safety margin. The modeled return constraint is conditional on accurate localization and the local graph containing a collision-free path to base. If no feasible exploration action exists, or if
% \begin{equation}
% D_i(x_{i,t},x_i^0;\mathcal{G}_{i,t})+\epsilon \geq B_{i,t},
% \end{equation}
% the robot switches to return mode and follows the shortest graph path back to base.

\begin{figure*}[t]
\vspace{0pt}
\captionsetup{skip=2pt}
\centering
\includegraphics[width=0.99\textwidth]{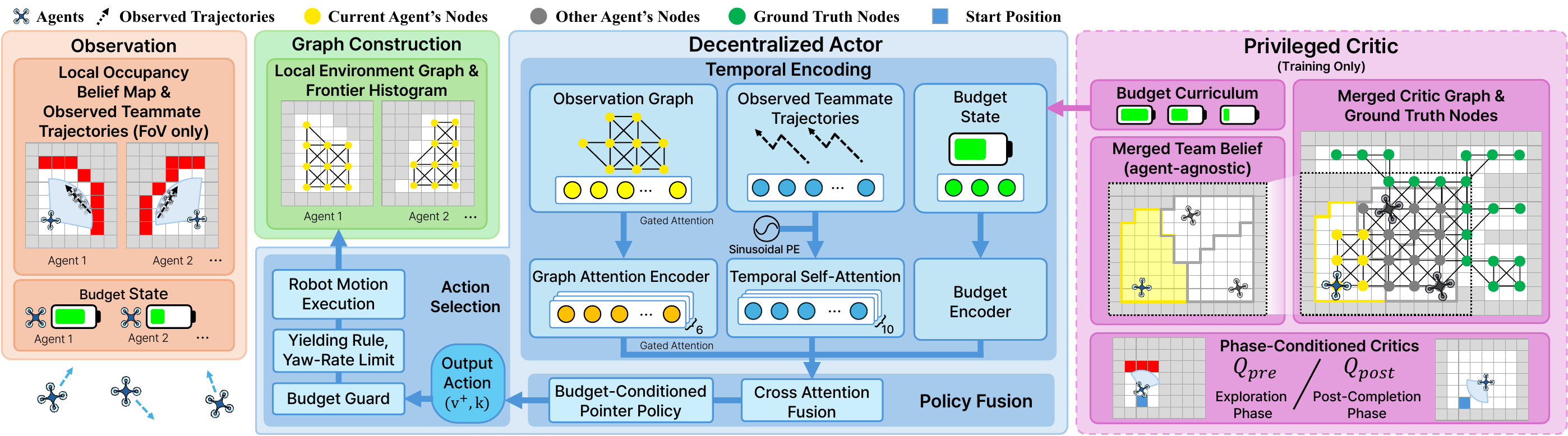}
\caption{Overall pipeline of Dec-MARVEL. Each robot builds a local graph from its observation and encodes visible teammate trajectories; the decentralized actor fuses them to output a waypoint-heading pair, and the training-only TOP critic augments value estimation with the decomposition \eqref{top_critic}.}
\label{fig:pipeline}
\vspace{-18pt}
\end{figure*}

\subsection{Map Beliefs and Coverage Objective}
\vspace{-3pt}

Let $M^*$ denote the ground-truth occupancy map, $c$ denote a grid cell and $M_{i,t}$ the local belief map of robot $i$. The merged team belief, used only for training supervision and evaluation, is
\begin{equation}
\label{merged_team_belief}
M_t^{team}(c)=
\begin{cases}
\mathrm{occupied}, & \exists i:\;M_{i,t}(c)=\mathrm{occupied},\\
\mathrm{free}, & \exists i:\;M_{i,t}(c)=\mathrm{free},\\
\mathrm{unknown}, & \mathrm{otherwise}.
\end{cases}
\end{equation}
Team exploration coverage $C_t$ is the fraction of ground-truth free cells marked free in the merged team belief,
$C_t = |\{c:M_t^{\mathrm{team}}(c)=\mathrm{free}\}| / |\{c:M^*(c)=\mathrm{free}\}|$.
An episode is successful when $C_t\geq C_{\mathrm{succ}}$, where we use $C_{\mathrm{succ}}=0.99$.

\subsection{Observation and Action Spaces}
\vspace{-3pt}
Each robot observes only deployment-time local information: graph structure, directional frontier, visitation history, budget state, and visible teammate trajectories. From reachable free space, it builds graph $\mathcal{G}_{i,t}=(V_{i,t},E_{i,t})$ using a $0.4$\,m occupancy grid and $4$\,m graph nodes. Each node $v$ has
\begin{equation}
\label{node_feature}
x_v=[\Delta x_v,\Delta y_v,u_v,g_v,o_v,\bar{\theta}_v,m_v,d_v^0,\nu_v],
\end{equation}
encoding relative position, frontier utility, guidepost\slash occu\-pancy state, best sensing heading, teammate-visit evidence, distance-to-base, and self-visitation.

To support budget generalization, we use log-normalized budget features:
\begin{equation}
\label{budget_feature}
b_{i,t}=\left[
\eta(B_i^0),\eta(B_{i,t}),\frac{B_{i,t}}{B_i^0},
\min\!\left(\frac{D_i(x_{i,t},x_i^0;\mathcal{G}_{i,t})}{\max(B_{i,t},1)},2\right)
\right],
\end{equation}
where $B_i^0$ is the initial budget, $B_{\max}$ is the largest training budget(Sec.~\ref{sec:mixture-based_budget_curriculum}), and $\eta(B)=\log(1+B)/\log(1+B_{\max})$ compresses the budget scale while preserving relative budget differences.

The action is defined as a waypoint--heading pair, $a_{i,t}=(v^+_{i,t},k_{i,t})$. $v^+_{i,t}$ is selected from $N(v_{i,t})$, a $5\times5$ local neighborhood centered at the current node with at most $K=25$ waypoint candidates, a size matched to the sensing range $d_s$. $k_{i,t}$ selects one of $H=3$ candidate headings drawn from a $36$-bin discretization of $360^\circ$ at $10^\circ$ per bin; we set $H=3$ empirically, as it provides directional flexibility without unnecessarily enlarging the action space, which contains at most $KH$ valid candidates after masking padding and self-slots, while zero-utility and previously visited nodes remain selectable for repositioning or return.

%$H=3$, $K_\theta=36$, $K = 5^2$, $KH=75$
% \footnote{The candidate headings are chosen }

\subsection{Reward and Objective Function}
\vspace{-3pt}

The reward encourages informative sensing, complementary coverage, and stable exploration, while return feasibility is enforced by the deterministic budget guard. For an active robot, the per-agent reward is
\begin{equation}
\label{eq:reward}
\begin{aligned}
r_{i,t}={}&
r^{\mathrm{FoV}}_{i,t}
+r^{\mathrm{merged}}_{i,t}
+r^{\mathrm{align}}_{i,t}
-0.15\,m_{i,t}\\
&-0.05\max(q_{i,t}-1,0)
+r^{\mathrm{team}}_t
+r^{\mathrm{indiv}}_{i,t}.
\end{aligned}
\end{equation}
% where $r_{i,t}^{\mathrm{FoV}}$ rewards visible-frontier coverage, $r_{i,t}^{\mathrm{merged}}$ rewards training-time merged-belief utility, and $r_{i,t}^{\mathrm{align}}$ rewards heading-motion alignment. The penalty terms discourage following recently observed teammate trails, encoded by $m_{i,t} = m_{v_{i,t}^+}$ and repeated low-utility moves through $q_{i,t}$, the number of consecutive waypoints with $0.5r^{\mathrm{FoV}}_{i,t}+r^{\mathrm{merged}}_{i,t} < 0.05$, with the count reset to zero on the first waypoint with $0.5r^{\mathrm{FoV}}_{i,t}+r^{\mathrm{merged}}_{i,t} \geq 0.05$. The team and individual completion bonuses are $5.0$ and $2.5$, respectively. Return feasibility is not rewarded but enforced by the deterministic budget guard.
\revision{The reward terms are computed using the executed waypoint and the heading realized after yaw-rate limiting. Let $n^{\mathrm{FoV}}_{i,t}$ denote the number of stored local frontiers visible from the executed waypoint within the executed FoV, and let $u^{\mathrm{team}}_t(v^+_{i,t})$ denote the waypoint's utility in the merged-map graph. Both quantities use the graph state before incorporating observations from the current transition. Observable frontiers satisfy a $9$\,m utility range and line-of-sight checks; node utilities with at most one observable frontier are set to zero. The merged utility counts frontiers across all directions and is zero if the waypoint is absent from the merged graph.

For the configured sensing range $d_s=10$\,m, FoV of $120^\circ$, and frontier resolution $h_f=0.8$\,m, define $Z=\lfloor 2d_s(\pi)/h_f\rfloor=78$ and $Z_{\mathrm{FoV}}=Z(120/360)=26$. The utility rewards are
\begin{equation}
\label{eq:utility_rewards}
r^{\mathrm{FoV}}_{i,t} = \frac{n^{\mathrm{FoV}}_{i,t}}{Z_{\mathrm{FoV}}}, \quad r^{\mathrm{merged}}_{i,t} = \frac{u^{\mathrm{team}}_t(v^+_{i,t})}{Z}.
\end{equation}

The alignment reward is
\begin{equation}
\label{eq:alignment_reward}
r^{\mathrm{align}}_{i,t} = \cos\!\left(\psi^{\mathrm{exec}}_{i,t}-\theta^{\mathrm{move}}_{i,t}\right),
\end{equation}
where $\psi^{\mathrm{exec}}_{i,t}$ is the executed sensing heading and $\theta^{\mathrm{move}}_{i,t}$ is the direction from the previous position to the executed waypoint, both in radians. For zero displacement, the implementation assigns $\theta^{\mathrm{move}}_{i,t}=0$.

Let $\mathcal{F}^{\mathrm{team}}_t$ denote the merged frontier set before the transition. The number of previously existing frontiers that disappear after the transition is
\begin{equation}
\label{eq:removed_frontiers}
d_t = \left|\mathcal{F}^{\mathrm{team}}_t \setminus \mathcal{F}^{\mathrm{team}}_{t+1}\right|.
\end{equation}
Newly created frontiers do not subtract from this count. Let $h_t^{\mathrm{team}}\in\{0,1\}$ indicate whether team completion has already occurred, and define the first-completion event as
\begin{equation}
\label{eq:team_completion_event}
e_t^{\mathrm{team}} = (1-h_t^{\mathrm{team}})\mathds{I}[C_{t+1}\geq C_{\mathrm{succ}}],
\end{equation}
where $C_{\mathrm{succ}}=0.99$. The team reward is
\begin{equation}
\label{eq:team_reward}
r^{\mathrm{team}}_t = (1-h_t^{\mathrm{team}})\left(\frac{d_t}{39}-0.5\right)+5e_t^{\mathrm{team}},
\end{equation}
where $39=\lfloor d_s(\pi)/h_f\rfloor$. The flag is updated as $h_{t+1}^{\mathrm{team}}=h_t^{\mathrm{team}}+e_t^{\mathrm{team}}$. The team reward is zero after completion and is added to each active explorer's reward.

Individual coverage is
\begin{equation}
\label{eq:individual_coverage}
C^{\mathrm{indiv}}_{i,t} = \frac{|\{c:M_{i,t}(c)=\mathrm{free}\}|}{|\{c:M^*(c)=\mathrm{free}\}|}.
\end{equation}

Let $h^{\mathrm{indiv}}_{i,t}\in\{0,1\}$ indicate whether robot $i$ has already received its individual completion bonus. Then
\begin{equation}
\label{eq:individual_reward}
r^{\mathrm{indiv}}_{i,t} = 2.5(1-h^{\mathrm{indiv}}_{i,t})\mathds{I}[C^{\mathrm{indiv}}_{i,t+1}\geq C_{\mathrm{succ}}].
\end{equation}
This bonus requires the robot's own map to reach $99\%$ coverage of ground-truth free space; returning to base is not required.

Finally, $m_{i,t}$ is the stored teammate-visit evidence at the executed waypoint, clipped to $[0,1]$. The counter $q_{i,t}$ records consecutive actions satisfying $0.5r^{\mathrm{FoV}}_{i,t}+r^{\mathrm{merged}}_{i,t}<0.05$ and resets to zero otherwise.}

The policy optimizes expected discounted team reward under the return-feasibility constraint:
\begin{equation}
\label{eq:constrained_objective}
\begin{split}
\max_\theta\;&
\mathbb{E}_{\tau\sim\pi_\theta}
\!\left[
\sum_{t=0}^{T}\gamma^t
\sum_{i=1}^{N} r_{i,t}
\right]\\
\mathrm{s.t.}\;&
\|x'_{i,t}-x_{i,t}\|_2
+
D_i(x'_{i,t},x_i^0;\mathcal{G}_{i,t})
+\epsilon
\le B_{i,t},
\ \forall i,t.
\end{split}
\end{equation}

\section{Method}
\vspace{-3pt}

\subsection{System Overview}
\vspace{-3pt}
% Fig.~\ref{fig:pipeline} illustrates the overall Dec-MARVEL pipeline. The Dec-MARVEL pipeline alternates between graph construction, budget-conditioned policy inference, continuous simulation of robot motion and sensor heading, belief-map updates, reward assignment, and off-policy critic learning. From raw observations, each robot constructs a local spatial graph and encodes observed teammate trajectories; the decentralized actor then fuses these representations through temporal encoding and policy fusion to select a waypoint-heading action. A privileged critic, available only during training, augments value estimation with merged-belief information from all agents. Each robot independently updates its local occupancy belief using FoV ray casting, extracts frontier nodes, refreshes its spatial graph, and evaluates the policy without receiving any teammate map, goal, action, or message. Candidate actions are filtered by return-budget feasibility and collision resolution before execution, and the final heading is constrained by yaw-rate limits to prevent unrealistically instantaneous sensor reorientation.
Dec-MARVEL approximately solves the constrained team objective in \eqref{eq:constrained_objective} with a decentralized policy trained under centralized supervision (Fig.~\ref{fig:pipeline}). \revision{At deployment, each robot's actor maps only its own belief map, local graph, remaining budget, and intermittently observed teammate trajectories to a waypoint-heading action, without exchanging maps, goals, or actions; the budget guard filters return-infeasible actions, and yaw-rate limits constrain the executed heading. During training, a task-oriented privileged critic uses merged team-belief information that is unavailable at deployment. A centralized arbiter resolves duplicate waypoint selections during training (Sec.~\ref{sec:soft_actor_critic}) and, because of the small arenas, in the hardware experiment (Sec.~\ref{sec:real-world_experiments}); in all other evaluations, each robot instead yields to visible teammates that are closer to its selected waypoint (Sec.~\mbox{\ref{sec:runtime_feasibility}}).}

\subsection{Graph and Frontier Abstraction}
\vspace{-3pt}
Each robot extracts frontiers from its local occupancy belief and maintains a quadtree-backed node graph. A node's utility is represented both as a scalar and as an angular frontier histogram with 36 bins. Observable frontiers within the $9$\,m utility range and line of sight are binned by angle, and a sliding FoV window identifies the $H$ candidate headings covering the largest frontier mass. If a node has no local frontier utility, candidate headings are generated from the $A^*$ guidepost path or neighboring graph directions, allowing the action space to remain well-defined during transit.

\subsection{Opportunistic Trajectory Observation as Coordination}
\label{sec:trajectory_observation}
\vspace{-3pt}
A robot updates its teammate-trajectory buffer only when another robot is simultaneously within sensing range, inside its FoV, and visible through line of sight. \revision{The main evaluation assumes accurate teammate detections and exact self-localization. Nearest-node association is unchanged when a position perturbation remains within the same node's Voronoi region, although continuous trajectory features can still change. Sec.~\ref{sec:ablation_robustness} evaluates position and heading perturbations, including changes in node association; identity errors are not evaluated.} If a teammate is not visible, the corresponding token is zeroed and masked rather than filled using privileged state. For each visible teammate $j$, the encoder receives the last $L=10$ observed states:
\begin{equation}
\label{eq:trajectory_encoding}
    \xi_{j,\ell} =
    [\Delta x_{j,\ell}, \Delta y_{j,\ell},
    \sin\psi_{j,\ell}, \cos\psi_{j,\ell}, \alpha_{j,\ell}],
\end{equation}
where $\Delta x_{j,\ell}$ and $\Delta y_{j,\ell}$ are relative coordinates in the observing robot's frame, $\psi_{j,\ell}$ is the observed heading, and $\alpha_{j,\ell}$ encodes age since observation. \revision{Trajectory tokens and teammate-visit evidence provide
the actor with coordination cues derived solely from local detections.}

\subsection{Runtime Feasibility}
\label{sec:runtime_feasibility}
\vspace{-3pt}
% \revision{After each robot independently proposes a waypoint, a centralized simulation process resolves duplicate target selections. For a shared target, the robot nearest to that target keeps it; each other robot is reassigned to an available waypoint nearest to its originally selected target. This process accesses the robots’ proposed waypoints and positions and is applied identically to Dec-MARVEL and every baseline. The waypoint-selection policy uses only local observations, but the complete simulated execution includes this centralized arbitration. The policy proposes actions, but execution is constrained by deterministic, training-free mechanisms. For each proposed waypoint $x'$, the budget guard evaluates the feasibility condition in \eqref{eq:budget_feasibility} and \eqref{eq:return_mode_change}. If the action violates the return constraint, the robot switches to return mode and follows the $A^\ast$ path back to base.}
\revision{At deployment, robot $i$ checks its selected waypoint only against currently visible teammates, whose positions are associated with nodes in its local graph. It yields when a visible teammate is closer to that waypoint and executes its next most probable return-feasible action. This rule uses neither teammate actions, communication, nor centralized arbitration. Unobserved or equal-distance conflicts may remain and are delegated to low-level collision avoidance. The budget guard enforces \eqref{eq:budget_feasibility}--\eqref{eq:return_mode_change} and switches the robot to its $A^\ast$ return path when necessary.}

\section{Decentralized Actor Architecture}

Like MARVEL~\cite{chiun2025marvel}, Dec-MARVEL uses a graph-attention actor to select waypoint-heading actions under FoV constraints, but its action selection is budget-conditioned and observed teammate trajectories are preserved as per-timestep tokens rather than pooled into a single vector. The actor consists of a graph-attention encoder over the node graph (Sec.~\ref{waypoint_heading_actor}), a temporal encoder over observed teammate trajectories (Sec.~\ref{temporal_trajectory_encoder}), and a budget-conditioned pointer decoder that fuses the two and selects an action (Sec.~\ref{action_decoding}).

\subsection{Budget-Aware Waypoint-Heading Actor}
\label{waypoint_heading_actor}
\vspace{-3pt}
Node features \eqref{node_feature} are projected to $d=128$ and encoded by six masked graph-attention layers with four attention heads.

\vspace{-10pt}
\begin{equation}\label{eq:graph_attention}
\begin{split}
A_m
&=
\mathrm{softmax}\!\left(
\frac{Q_mK_m^\top}{\sqrt{d_k}}
+\mathcal{M}
\right)V_m,\\
\tilde z_{v,m}
&=
\sigma(q_v^\top w_m^g)[A_m]_v .
\end{split}
\end{equation}

% \begin{equation}\label{eq:graph_attention2}
% \begin{split}
% \mathrm{head}_{v,m}
% &=
% \left[\mathrm{softmax}\!\left(
% \frac{Q_mK_m^\top}{\sqrt{d_k}}
% +\mathcal{M}
% \right)V_m\right]_v,\\
% \tilde z_{v,m}
% &=
% \sigma(q_v^\top w_m^g)\mathrm{head}_{v,m},
% \end{split}
% \end{equation}

Here, $\mathcal{M}$ masks invalid or non-adjacent nodes, and $\sigma(q_v^\top w^g_m)$ gates head $m$ using the query node, giving the per-head output $\tilde z_{v,m}$ in \eqref{eq:graph_attention}. The gated heads are concatenated over $m$, projected, and passed through residual feed-forward blocks with layer normalization to form the per-node graph embedding $z^g_v$. Directional frontier histograms are encoded as $e^f_v = \mathrm{Conv1D}(f_v)$ and fused with it as $z_v = W_f[z^g_v;e^f_v]$.

\subsection{Temporal Trajectory Encoder}
\label{temporal_trajectory_encoder}
\vspace{-3pt}
The trajectory tokens $\xi_{j,\ell}$ defined in \eqref{eq:trajectory_encoding} are encoded using two masked temporal-attention layers with four heads and $64$-dimensional embeddings. Each observation is augmented with sinusoidal temporal positional encoding
% \begin{equation}\label{eq:trajectory_attention}
   $ e_\ell = W_\xi \xi_{j,\ell} + p_\ell $
% \end{equation}
where $p_\ell$ encodes the timestep index, following Motion Transformer model~\cite{shi2022motion}. The valid agent-time tokens $\tau_{j,\ell}$ are kept rather than pooled. When node correspondence is available, each token is fused with the nearest observed graph node:
\begin{equation}\label{eq:token_node_fusion}
    \hat{\tau}_{j,\ell}=W_\tau[W_p\tau_{j,\ell};W_nz_{\mathrm{node}(j,\ell)}].
\end{equation}
Candidate actions cross-attend to the fused tokens $\hat{\tau}_{j,\ell}$ in \eqref{eq:token_node_fusion}, so each waypoint-heading candidate is evaluated in the context of recently observed teammate motion.

\subsection{Budget-Conditioned Action Decoding}
\label{action_decoding}
\vspace{-3pt}
The policy context $\zeta_{i,t}$ combines the current-node embedding $z_{v_i}$, and the remaining-budget features $b_{i,t}$ in \eqref{budget_feature}:
\begin{equation*}\label{eq:budget_conditioned_decoding}
    \zeta_{i,t}=W_\zeta[z_{v_i};W_bb_{i,t}].
\end{equation*}
For each neighbor-heading candidate $a=(v',k)$, the action embedding is
\begin{equation*}\label{eq:action_embedding}
    \rho_a=W_a[z_{v'};W_{h}h_{v'};W_\theta\theta(v',k)],
\end{equation*}
where $h_{v'}$ embeds the heading(s) from which node $v'$ has already been sensed, and $\theta(v',k)$ is the binary heading-window vector for the candidate heading $k$. After cross-attending to the fused trajectory tokens $\hat{\tau}_{j,\ell}$ of \eqref{eq:token_node_fusion}, a pointer head produces masked log-probabilities:

\begin{equation}\label{eq:pointer_policy}
\log \pi_\theta(a|o,b)=
\log\left[
\operatorname{softmax}_{a'}
\left(
\frac{\zeta^\top W_q^\top W_k\rho_{a'}}{\sqrt d}
\right)
\right]_a.
\end{equation}
Actions are sampled during training and selected greedily at evaluation; invalid graph actions are masked.

% \subsection{Kinematic Heading Realization}
% \vspace{-3pt}

% The policy selects a desired heading bin, but the simulator applies a yaw-rate-constrained heading update. Given current heading $\psi_t$, desired heading $\psi^*$, travel distance $d_t$, velocity $v$, and yaw-rate limit $\omega_\textit{max}$, the achievable heading change is bounded by $(d_t/v)\omega_\textit{max}$. This prevents the learned policy from exploiting unrealistically instantaneous sensor reorientation and ensures that heading decisions reflect physically realizable sensor configurations.

\section{Centralized Training}
\subsection{Soft Actor Critic}
\label{sec:soft_actor_critic}
Dec-MARVEL uses discrete Soft Actor-Critic (SAC)~\cite{haarnoja2018soft} with twin Q-critics. For each transition $(s,a,r,s',d)$, where $d$ indicates termination, we normalize $r$ using moving estimates of mean and variance (update rate $0.001$, initialized to zero and one): $\hat r=(r-\mu)/\max(\sqrt{v},10^{-6})$. The discrete-SAC target averages over next actions under $\pi_\theta$, uses the smaller target-critic value and an entropy term, and discounts the next-state value by $\gamma(1-d)$. Each critic minimizes squared Bellman error. The pointer policy in~\eqref{eq:pointer_policy} and temperature $\alpha$ use the discrete-SAC objectives. We set target entropy to $0.5\log|A|$, clamp $\log\alpha$ to $[-4,2]$, and update target critics by Polyak averaging with $\tau=0.003$. \revision{During training, the simulator resolves duplicate waypoint selections centrally: for a contested waypoint, the closest robot keeps it, and each of the others executes its next most probable action with a different waypoint. This arbiter provides no input to the actor and, except in the hardware experiment (Sec.~\ref{sec:real-world_experiments}), is not used in the evaluations; the replay buffer stores the executed actions.}

\subsection{Phase-Conditioned Critics}
Exploration before and after mission completion forms two qualitatively different behavioral regimes. Near the completion boundary, target values change abruptly as the team reward structure switches, producing unstable Bellman targets under a single critic representation. Dec-MARVEL therefore equips each twin critic with two phase-conditioned Q-heads, $Q^{\mathrm{pre}}_{\phi_j}$ and $Q^{\mathrm{post}}_{\phi_j}$.
Given a phase indicator $\mathds{I}_{t}^{\mathrm{post}}=\mathds{I}[C_t\geq C_{\mathrm{succ}}]$, the active critic output becomes
\begin{equation}\label{eq:phase_conditioned_critic}
Q_{\phi_j}(s,a,\mathds{I}_t^{\mathrm{post}})
=(1-\mathds{I}_t^{\mathrm{post}})\,Q^{\mathrm{pre}}_{\phi_j}(s,a)
+\mathds{I}_t^{\mathrm{post}}\,Q^{\mathrm{post}}_{\phi_j}(s,a).
\end{equation}

This preserves shared representation learning while allowing phase-specific value geometry. The post-phase head is initialized from the pre-phase head for smoother optimization, and replay sampling becomes phase-stratified once sufficient post-completion samples are available.

\subsection{Task-Oriented Privileged (TOP) Critic}
\label{sec:top_critic}
The TOP critic closes the information gap between the decentralized actor and the team-level exploration objective during training. Rather than the agent-agnostic merge in \eqref{merged_team_belief}, TOP gives only the critic a task-specific privileged decomposition of map coverage: what the current robot has uncovered, what teammates have uncovered but remains unknown locally, and what ground-truth free space is unexplored by the team:
\begin{equation}\label{top_critic}
E_{i,t}(c)=
\begin{cases}
\mathrm{self}, & M_{i,t}(c)\neq u,\\
\mathrm{team}, & M_{i,t}(c)=u,\ \exists j\neq i:\ M_{j,t}(c)\neq u,\\
\mathrm{unexp}, & M^*(c)=f,\ \forall j:\ M_{j,t}(c)=u,
\end{cases}
\end{equation}
where $u$ and $f$ denote unknown and free cells, respectively. Aggregating these labels onto the privileged graph exposes team progress and remaining free space to the critic, aligning value targets with the exploration objective.
% For robot $i$, the privileged critic labels each map cell $c$ by its exposure source:
% \begin{equation}
% E_{i,t}(c)\in\{\mathrm{self},\mathrm{team},\mathrm{unk}\}
% \end{equation}
% \begin{equation}
% e_{v,i,t}^{r}
% =
% |\mathcal{C}(v)|^{-1}
% \sum_{c\in\mathcal{C}(v)}
% \mathbf{1}\{E_{i,t}(c)=r\},
% \quad r\in\{\mathrm{self},\mathrm{team},\mathrm{unk}\}.
% \end{equation}
% where $E_{i,t}(c)$ labels cell $c$ as observed by robot $i$, observed only by teammates, or still unknown to all robots.

% Several implementation choices further stabilize training. Target networks are updated using Polyak averaging with $\tau=0.003$, policy and critic gradients are clipped independently, and replay sampling is phase-stratified once enough post-completion transitions exist. Together, these mechanisms reduce temporal-difference spikes caused by frontier completion rewards and discrete phase transitions.

\section{Curriculum Learning Framework}
\label{sec:curriculum}

\subsection{Mixture-Based Budget Curriculum}
\label{sec:mixture-based_budget_curriculum}
\vspace{-3pt}
Dec-MARVEL adapts the target travel budget to recent strict mission success. Let $\bar{s}_k$ denote the strict mission-success rate over the most recent window of 8 episodes, and let $s_k$ be its exponential moving average, updated as $s_{k+1}=(1-\beta)s_k+\beta\bar{s}_k$ with $\beta=0.002$. Success tracking begins only after a warm-up of 1500 completed episodes; during warm-up we hold $s_k=0$, so the curriculum starts at the largest budget $B_\mathrm{start}$ and tightens as success accumulates. The target budget is
\begin{equation}
\label{eq:budget_curriculum}
B_k^{\mathrm{target}}
=
B_{\mathrm{end}}
+(B_{\mathrm{start}}-B_{\mathrm{end}})(1-s_k),
\end{equation}
where $B_{\mathrm{start}}=1024$\,m and $B_{\mathrm{end}}=640$\,m.

\subsection{Mixture Sampling for Generalization}
\vspace{-3pt}
Pure self-paced curricula can overfit to the current difficulty band, causing forgetting at easy or hard budgets. Dec-MARVEL therefore samples initial budgets from a mixture:
\begin{equation}\label{eq:mixture_sampling}
B^0_i \sim \begin{cases} 
    U(B_{\mathrm{end}}, B_{\mathrm{start}}) & \mathrm{with~prob.} p_u,\\
    \mathrm{clip}(B^{\mathrm{target}}_k + \delta,B_{\mathrm{end}}, B_{\mathrm{start}}) & \mathrm{otherwise},
\end{cases}
\end{equation}

with $p_u = 0.6$ and $\delta \sim \mathcal{U}(-128,128)$\,m. The uniform component maintains exposure to the full budget range, while target-centered sampling concentrates learning on the frontier of competence, supporting generalization to unseen budgets.

\begin{figure}[t]
\captionsetup{skip=2pt}
\centering
\includegraphics[width=0.9\columnwidth]{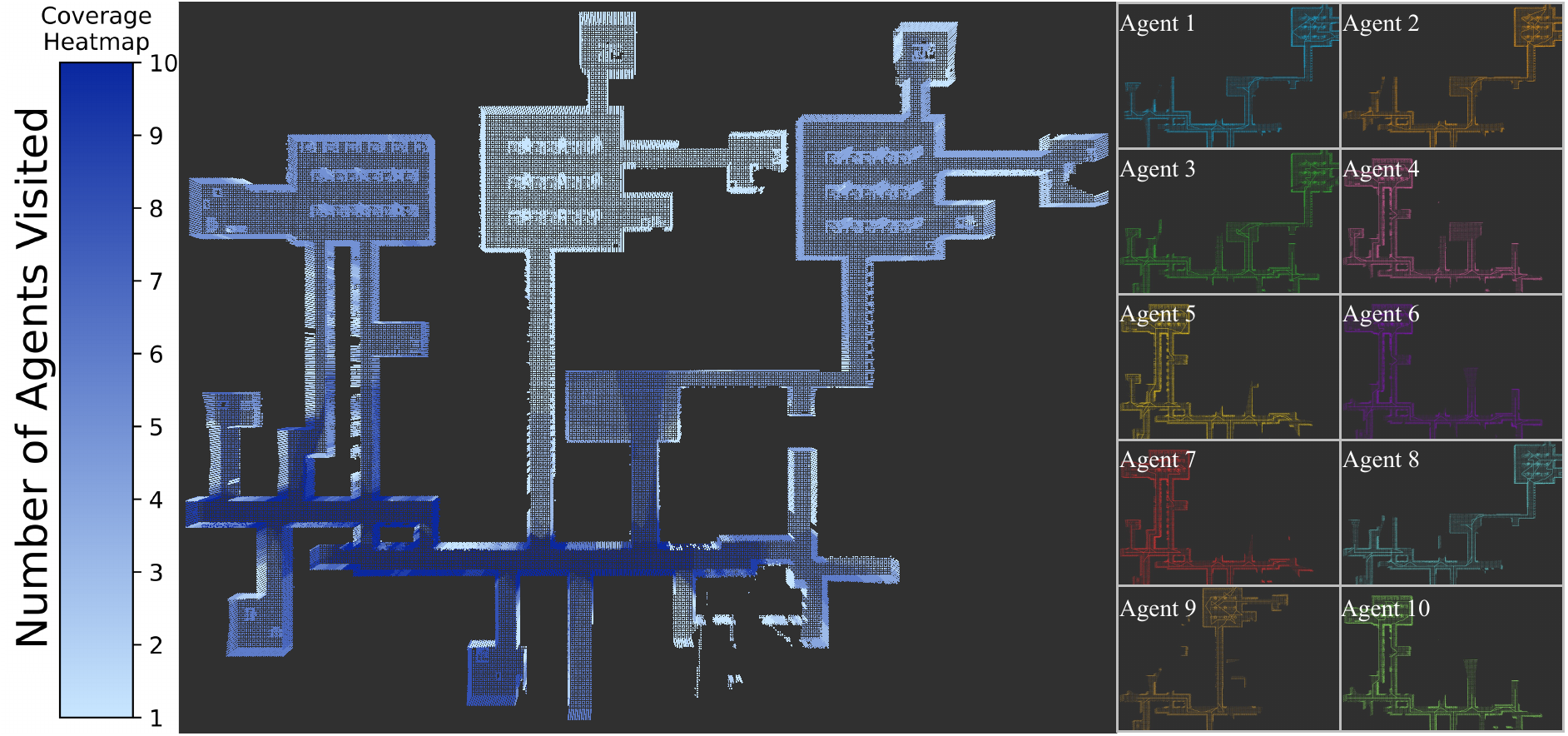}
\caption{3D simulation with 10 robots in the CMU Autonomous Exploration Development Environment~\cite{cao2022autonomous}. Left: accumulated sensing map, where color intensity marks sensing overlap. Right: per-agent sensing footprints, showing complementary coverage from opportunistic coordination.}
\label{fig:sensing_pointcloud}
\vspace{-18pt}
\end{figure}

\section{EXPERIMENTS}
\begin{table*}[tb]
\caption{\revision{\textbf{Quantitative comparison across budgets and team sizes.} We report the average exploration rate, success rate (percentage of trials reaching 99\% merged coverage), distance in meters to 99\% merged coverage (avg.$\pm$std.\ over the trials that reach it), and FoV overlap ratio (avg.$\pm$std.) for all methods, team sizes, and budgets, over 100 trials per configuration with a 120$^{\circ}$ FoV and sensor range $d_s = 10$\,m. $^\dagger$Adapted to the communication-free setting (Sec.~\ref{sec:baselines}). The downward and upward arrows indicate that lower and higher values are better, respectively; the best value in each row is in bold.}}
\vspace{-0.2cm}
\label{tab:main_results}
\centering
\resizebox{\textwidth}{!}{%
\tiny
\aboverulesep=0.2mm \belowrulesep=0.2mm
\begin{tabular}{c|c|c|c|c|c|c|c}
\toprule
Budget (m) & $N$ & Metrics & Nearest \cite{yamauchi1998frontier} & SMMR-Explore$^\dagger$ \cite{yu2021smmr} & NBVP \cite{bircher2016receding} & MARVEL$^\dagger$ \cite{chiun2025marvel} & Dec-MARVEL \\
\midrule
\multirow{12}{*}{720} & \multirow{4}{*}{$N\!=\!2$} & \textit{Avg.\ Exp.\ Rate $\uparrow$} & 0.769 & 0.842 & 0.809 & 0.880 & \textbf{0.945} \\
 &  & \textit{Success Rate (\%) $\uparrow$} & 7.0 & 9.0 & 8.0 & 17.0 & \textbf{46.0} \\
 &  & \textit{Dist.\ to 99\% (m) $\downarrow$} & 576.15 $\pm$ 60.59 & 540.63 $\pm$ 52.64 & 556.05 $\pm$ 43.32 & 536.20 $\pm$ 70.71 & \textbf{487.63} $\pm$ 102.67 \\
 &  & \textit{Overlap Ratio $\downarrow$} & 0.687 $\pm$ 0.245  & 0.595 $\pm$ 0.221 & 0.586 $\pm$ 0.283 & 0.799 $\pm$ 0.248 & \textbf{0.549} $\pm$ 0.299 \\
\cline{2-8}
 & \multirow{4}{*}{$N\!=\!4$} & \textit{Avg.\ Exp.\ Rate $\uparrow$} & 0.865 & 0.960 & 0.923 & 0.913 & \textbf{0.989} \\
 &  & \textit{Success Rate (\%) $\uparrow$} & 13.0 & 57.0 & 41.0 & 35.0 & \textbf{84.0} \\
 &  & \textit{Dist.\ to 99\% (m) $\downarrow$} & 492.32 $\pm$ 84.88 & 469.28 $\pm$ 87.00 & 441.62 $\pm$ 77.69  & 477.66 $\pm$ 83.13 & \textbf{379.41} $\pm$ 90.79 \\
 &  & \textit{Overlap Ratio $\downarrow$} & 0.793 $\pm$ 0.147 & 0.692 $\pm$ 0.158 & 0.742 $\pm$ 0.169 & 0.859 $\pm$ 0.153 & \textbf{0.688} $\pm$ 0.159 \\
\cline{2-8}
 & \multirow{4}{*}{$N\!=\!8$} & \textit{Avg.\ Exp.\ Rate $\uparrow$} & 0.937 & 0.991 & 0.983 & 0.938 & \textbf{0.997} \\
 &  & \textit{Success Rate (\%) $\uparrow$} & 36.0 & 85.0 & 85.0 & 46.0 & \textbf{94.0} \\
 &  & \textit{Dist.\ to 99\% (m) $\downarrow$} & 473.96 $\pm$ 92.63 & 406.65 $\pm$ 85.92 & 397.81 $\pm$ 92.10 & 462.38 $\pm$ 81.99 & \textbf{344.83} $\pm$ 95.41 \\
 &  & \textit{Overlap Ratio $\downarrow$} & 0.853 $\pm$ 0.118 & \textbf{0.792} $\pm$ 0.120 & 0.801 $\pm$ 0.115 & 0.906 $\pm$ 0.113 & 0.848 $\pm$ 0.114 \\
\midrule
\multirow{12}{*}{800} & \multirow{4}{*}{$N\!=\!2$} & \textit{Avg.\ Exp.\ Rate $\uparrow$} & 0.823 & 0.884 & 0.854 & 0.920 & \textbf{0.962} \\
 &  & \textit{Success Rate (\%) $\uparrow$} & 8.0 & 20.0 & 20.0 & 31.0 & \textbf{67.0} \\
 &  & \textit{Dist.\ to 99\% (m) $\downarrow$} & 581.74 $\pm$ 63.52 & 605.89 $\pm$ 74.10 & 617.76 $\pm$ 63.69 & 597.06 $\pm$ 90.10 & \textbf{531.83} $\pm$ 117.21 \\
 &  & \textit{Overlap Ratio $\downarrow$} & 0.692 $\pm$ 0.239 & 0.608 $\pm$ 0.228 & 0.592 $\pm$ 0.288 & 0.737 $\pm$ 0.260 & \textbf{0.561} $\pm$ 0.297 \\
\cline{2-8}
 & \multirow{4}{*}{$N\!=\!4$} & \textit{Avg.\ Exp.\ Rate $\uparrow$} & 0.910 & 0.978 & 0.948 & 0.956 & \textbf{0.996} \\
 &  & \textit{Success Rate (\%) $\uparrow$} & 23.0 & 68.0 & 56.0 & 53.0 & \textbf{92.0} \\
 &  & \textit{Dist.\ to 99\% (m) $\downarrow$} & 560.63 $\pm$ 112.62 & 498.53 $\pm$ 104.09 & 490.55 $\pm$ 108.13 & 525.24 $\pm$ 106.49 & \textbf{403.56} $\pm$ 113.49 \\
 &  & \textit{Overlap Ratio $\downarrow$} & 0.790 $\pm$ 0.147 & 0.693 $\pm$ 0.159 & 0.739 $\pm$ 0.168 & 0.811 $\pm$ 0.160 & \textbf{0.689} $\pm$ 0.161 \\
\cline{2-8}
 & \multirow{4}{*}{$N\!=\!8$} & \textit{Avg.\ Exp.\ Rate $\uparrow$} & 0.963 & 0.997 & 0.991 & 0.972 & \textbf{0.999} \\
 &  & \textit{Success Rate (\%) $\uparrow$} & 56.0 & 93.0 & 90.0 & 71.0 & \textbf{98.0} \\
 &  & \textit{Dist.\ to 99\% (m) $\downarrow$} & 527.65 $\pm$ 108.73 & 423.69 $\pm$ 100.33 & 407.90 $\pm$ 99.10 & 484.63 $\pm$ 125.48 & \textbf{353.63} $\pm$ 111.01 \\
 &  & \textit{Overlap Ratio $\downarrow$} & 0.853 $\pm$ 0.120 & \textbf{0.792} $\pm$ 0.120 & 0.800 $\pm$ 0.114 & 0.855 $\pm$ 0.136 & 0.845 $\pm$ 0.110 \\
\midrule
\multirow{12}{*}{1024} & \multirow{4}{*}{$N\!=\!2$} & \textit{Avg.\ Exp.\ Rate $\uparrow$} & 0.938 & 0.965 & 0.940 & 0.973 & \textbf{0.986} \\
 &  & \textit{Success Rate (\%) $\uparrow$} & 38.0 & 62.0 & 51.0 & \textbf{87.0} & \textbf{87.0} \\
 &  & \textit{Dist.\ to 99\% (m) $\downarrow$} & 769.59 $\pm$ 117.94 & 726.13 $\pm$ 117.81 & 727.67 $\pm$ 123.99 & 730.08 $\pm$ 134.79 & \textbf{592.06} $\pm$ 149.17 \\
 &  & \textit{Overlap Ratio $\downarrow$} & 0.702 $\pm$ 0.239 & 0.610 $\pm$ 0.231 & 0.589 $\pm$ 0.287 & 0.755 $\pm$ 0.289 & \textbf{0.554} $\pm$ 0.287 \\
\cline{2-8}
 & \multirow{4}{*}{$N\!=\!4$} & \textit{Avg.\ Exp.\ Rate $\uparrow$} & 0.977 & 0.998 & 0.988 & 0.971 & \textbf{0.999} \\
 &  & \textit{Success Rate (\%) $\uparrow$} & 70.0 & 95.0 & 85.0 & 88.0 & \textbf{99.0} \\
 &  & \textit{Dist.\ to 99\% (m) $\downarrow$} & 705.74 $\pm$ 136.03 & 573.06 $\pm$ 155.49 & 574.78 $\pm$ 166.39 & 697.05 $\pm$ 146.13 & \textbf{429.39} $\pm$ 142.77 \\
 &  & \textit{Overlap Ratio $\downarrow$} & 0.794 $\pm$ 0.149 & \textbf{0.691} $\pm$ 0.157 & 0.731 $\pm$ 0.161 & 0.870 $\pm$ 0.156 & 0.695 $\pm$ 0.160 \\
\cline{2-8}
 & \multirow{4}{*}{$N\!=\!8$} & \textit{Avg.\ Exp.\ Rate $\uparrow$} & 0.993 & \textbf{1.000} & 0.999 & 0.975 & \textbf{1.000} \\
 &  & \textit{Success Rate (\%) $\uparrow$} & 87.0 & \textbf{100.0} & 98.0 & 92.0 & \textbf{100.0} \\
 &  & \textit{Dist.\ to 99\% (m) $\downarrow$} & 603.94 $\pm$ 141.14 & 445.11 $\pm$ 125.29 & 428.74 $\pm$ 121.12 & 613.14 $\pm$ 176.28 & \textbf{379.25} $\pm$ 135.15 \\
 &  & \textit{Overlap Ratio $\downarrow$} & 0.855 $\pm$ 0.120 & \textbf{0.792} $\pm$ 0.120 & 0.800 $\pm$ 0.113 & 0.884 $\pm$ 0.128 & 0.868 $\pm$ 0.106 \\
\bottomrule
\end{tabular}%
}
\vspace{-0.5cm}
\end{table*}
\subsection{Experimental Setup}
\vspace{-3pt}
We evaluate Dec-MARVEL in procedurally generated 2D indoor maps with directional sensing ($120^\circ$ FoV, $d_s=10$\,m); a representative multi-robot simulation is shown in Fig.~\ref{fig:sensing_pointcloud}. We test three travel budgets, $B$ in $\{720,800,1024\}$\,m, and three team sizes, $N\in\{2,4,8\}$, yielding nine budget--team configurations. We use 100 held-out scenarios per configuration, giving 900 unique evaluation scenarios; every method is evaluated on the same scenarios. \revision{Robots start from distinct nodes within a common base region.} We report merged exploration rate, success rate at $99\%$ merged coverage, maximum per-robot distance to reach $99\%$ merged coverage \revision{(averaged over the trials that reach it)}, and FoV overlap ratio. \revision{We define the FoV overlap ratio as the fraction of the area sensed by the team that has been sensed by more than one robot, measured when the merged coverage first reaches 90\%, or at the end of the episode if it never does.}
% NOTE: definition rewritten to match the CSV column overlap_until_0_90 (cumulative: until90 <= until99 <= episode in every row); confirm against the evaluation code. Percentage improvements are computed against the strongest baseline for each metric and configuration.

% NOTE (alignment check, 2026-10-02, 23:45): the Dec-MARVEL row now matches the other tables: merged distance 403.56 +- 113.49 (Tables I, III),
% overlap 0.689 +- 0.161 (Tables I, IV; was 0.691), individual distance 604.04 +- 74.66 (Table III).
% The "w/o TOP Critic" row was filled with the new result on 2026-10-03 (501.83 +- 123.27; overlap 0.873 +- 0.151). The percentages in
% Sec. VIII-D (+24.4%, 7.3%, 4.3%) are relative to the Dec-MARVEL row (403.56 m); update them whenever this table changes.
\begin{table}[tb]
\caption{Ablation study of Dec-MARVEL at $B=800$\,m with $N=4$ agents.
\vspace{-0.2cm}
% We evaluate the contribution of curriculum learning, merged critic supervision, and dual-head critic design.
Each entry reports the average value with standard deviation.}
\label{tab:ablation}
\centering
\resizebox{\columnwidth}{!}{%
\footnotesize
\begin{tabular}{>{\centering\arraybackslash}m{2.5cm}
                >{\centering\arraybackslash}m{2cm}
                >{\centering\arraybackslash}m{2cm}
                >{\centering\arraybackslash}m{1.5cm}}
\toprule
Method
& \shortstack{Dist to 99\%\\Merged $\downarrow$}
& \shortstack{Dist to 99\%\\Individual $\downarrow$}
& Overlap $\downarrow$ \\
\midrule

% training in progress
\shortstack{Dec-MARVEL w/o\\TOP Critic}
& 501.83  $\pm$ 123.27
& Did not reach 99\%
& 0.873  $\pm$ 0.151 \\
\cmidrule(lr){1-4}

\shortstack{Dec-MARVEL w/o\\Mixture-based \\Curriculum}
& 432.84  $\pm$ 121.67
& 679.18  $\pm$ 56.86
& 0.749  $\pm$ 0.181 \\
\cmidrule(lr){1-4}

\shortstack{Dec-MARVEL w/o\\Phase-Conditioned Critic}
& 421.11  $\pm$ 105.99
& 607.20  $\pm$ 74.85
& 0.745  $\pm$ 0.169 \\
\cmidrule(lr){1-4}

Dec-MARVEL
& \textbf{403.56}  $\pm$ 113.49
& \textbf{604.04} $\pm$ 74.66
& \textbf{0.689}  $\pm$ 0.161 \\

\bottomrule
\end{tabular}}
\vspace{-20pt}
\end{table}

\subsection{Baselines}
\label{sec:baselines}
\vspace{-3pt}
\revision{All methods plan from local maps without map exchange; teammate positions are available only within sensing range, FoV, and line of sight. We compare against four baselines. \textbf{Nearest Frontier}~\cite{yamauchi1998frontier} selects the closest local frontier. \textbf{SMMR-Explore (adapted)}~\cite{yu2021smmr} retains multi-target potential-field planning but replaces exchanged submaps with each robot's local map and uses only visible teammates for repulsion. \textbf{NBVP}~\cite{bircher2016receding} replans next-best views using single-step information gain. \textbf{MARVEL (adapted)}~\cite{chiun2025marvel} replaces shared-map inputs with local maps and is retrained under the limited-budget setting; its actor receives teammate information during training, but none is provided at test time. It lacks Dec-MARVEL's budget-conditioned inputs, phase-conditioned critics, and mixture-based curriculum. The SMMR-Explore and MARVEL results therefore represent communication-free adaptations, not their original full systems.}

\revision{During simulation testing, every method uses the same budget guard and return-to-base procedure, which also apply to baselines whose planners do not condition on remaining budget. No centralized arbitration is used; the yielding rule of Sec.~\ref{sec:runtime_feasibility} is part of Dec-MARVEL and is not applied to the baselines.}

\begin{table}[tb]
\caption{Robustness of a four-robot team to robot failures across exploration budgets.}
\vspace{-0.2cm}
\label{tab:breakdown_budget}
\centering
\resizebox{\columnwidth}{!}{%
\footnotesize
\begin{tabular}{>{\centering\arraybackslash}m{1cm}
                >{\centering\arraybackslash}m{1cm}
                >{\centering\arraybackslash}m{1.5cm}
                >{\centering\arraybackslash}m{1.2cm}
                >{\centering\arraybackslash}m{2cm}
                >{\centering\arraybackslash}m{2cm}}
\toprule
Budget (m)
& \shortstack{Broken\\Agents}
& \shortstack{Avg.\ Exp.\\ Rate $\uparrow$}
& \shortstack{Success\\ (\%) $\uparrow$}
& \shortstack{Merged 99\% $\downarrow$}
& \shortstack{Individual\\99\% $\downarrow$} \\
% & Overlap $\downarrow$ \\
\midrule

% \multirow{3}{*}{70}
% & 2
% & 0.9116
% & 52.0
% & 355.908  $\pm$ 78.588)
% & N/A
% & 0.096  $\pm$ 0.142) \\

% & 1
% & 0.9505
% & 59.0
% & 355.067  $\pm$ 68.173)
% & N/A
% & 0.096  $\pm$ 0.136) \\

% & 0
% & \textbf{0.9780}
% & \textbf{81.0}
% & \textbf{336.742}  $\pm$ 84.722)
% & N/A
% & \textbf{0.088}  $\pm$ 0.136) \\

% \midrule

% \multirow{3}{*}{80}
% & 2
% & 0.9189
% & 62.0
% & 370.901  $\pm$ 91.491)
% & 587.559  $\pm$ 0.000)
% & 0.084  $\pm$ 0.131) \\

% & 1
% & 0.9809
% & 85.0
% & 393.982  $\pm$ 91.099)
% & 599.380  $\pm$ 0.000)
% & 0.084  $\pm$ 0.127) \\

% & 0
% & \textbf{0.9926}
% & \textbf{92.0}
% & \textbf{357.980}  $\pm$ 83.571)
% & 584.401  $\pm$ 15.245)
% & \textbf{0.079}  $\pm$ 0.131) \\

% \midrule

\multirow{3}{*}{720}
& 2
& 0.961
& 78.0
& 389.34 $\pm$ 75.01
& 620.51 $\pm$ 37.80 \\
% & 0.080  $\pm$ 0.129) \\

& 1
& 0.977
& 76.0
& 388.63 $\pm$ 84.03
& 616.48 $\pm$ 53.80 \\
% & 0.081  $\pm$ 0.127) \\

& 0
& \textbf{0.989}
& \textbf{84.0}
& \textbf{379.41} $\pm$ 90.79
& \textbf{613.07} $\pm$ 43.19 \\
% & \textbf{0.083}  $\pm$ 0.128) \\

\midrule

\multirow{3}{*}{800}
& 2
& 0.980
& 85.0
& 406.55 $\pm$ 87.18
& 697.73 $\pm$ 56.12 \\
% & 0.076  $\pm$ 0.127) \\

& 1
& 0.990
& 91.0
& 409.32 $\pm$ 114.04
& 699.03 $\pm$ 53.96 \\
% & 0.078  $\pm$ 0.123) \\

& 0
& \textbf{0.996}
& \textbf{92.0}
& \textbf{403.56} $\pm$ 113.49
& \textbf{604.04} $\pm$ 74.66 \\
% & \textbf{0.070}  $\pm$ 0.125) \\

\midrule

\multirow{3}{*}{1024}
& 2
& 0.985
& 91.0
& 474.79 $\pm$ 115.53
& 775.23 $\pm$ 85.56 \\
% & 0.074  $\pm$ 0.123) \\

& 1
& 0.996
& 96.0
& 442.26 $\pm$ 125.41
& 783.98 $\pm$ 84.36 \\
% & 0.073  $\pm$ 0.120) \\

& 0
& \textbf{0.999}
& \textbf{99.0}
& \textbf{429.39} $\pm$ 142.77
& \textbf{689.30} $\pm$ 85.85 \\
% & \textbf{0.072}  $\pm$ 0.119) \\

\bottomrule
\end{tabular}}
\vspace{-4pt}
\end{table}

% \subsection{Ablation Studies}
% \vspace{-3pt}
% Ablation studies will evaluate: trajectory encoder (no trajectory, pooled trajectory vs.\ full token cross-attention), budget features (no budget features, no runtime guard), phase-split critic heads (single head vs.\ phase-conditioned), merged-belief critic (local critic only vs.\ privileged merged-belief), gated attention (ungated vs.\ gated), reward shaping terms (trail penalty, alignment reward), and curriculum mixture sampling (pure self-paced vs.\ mixture).
% NOTE (alignment check, 2026-10-02, 23:30): the Nominal rows match Table I at B=800 m (0.962/67/0.561, 0.996/92/0.689, 0.999/98/0.845).
% The rho, Explore, and Success values of the Drop/Range rows (and rho of the Nominal rows) are identical to revision 3, i.e., to the
% previous Dec-MARVEL run; only the Nominal Explore/Success and the Overlap column changed. Confirm that these rows come from the final policy.
% Overlap std of the Nominal rows = Table I (800 m), three decimals: 0.297, 0.161, 0.110 (author's decision, 2026-10-03; were 0.24, 0.19, 0.11).
% The authors replaced the stds of the Drop/Range rows by three-decimal values and added the Range r = 7.5 m and 5 m rows on 2026-10-03.
% The authors confirmed on 2026-10-04 that all values of this table are correct, including N=8 Drop p=.75 (+-0.069),
% N=2 Range r=7.5 m (+-0.297, equal to the Nominal row), and the rho values of the Range rows.
\begin{table}[t]
\caption{\revision{Dec-MARVEL under sparse teammate observations
($B=800$\,m; 100 trials per condition). Map sensing remains
at 10\,m. Here $p$ is detection-dropout probability,
$r$ is teammate-detection range, and $\rho$ is the fraction
of exploration-mode decision steps with at least one teammate observed. Overlap is reported as avg.$\pm$std.}}
\label{tab:visibility_robustness}
\centering
% Same font size as Table III: Table III is scaled by \columnwidth / (8.7cm + 12\tabcolsep) = 245.72pt / 319.54pt = 0.769.
\scalebox{0.8}{%
\footnotesize
\setlength{\tabcolsep}{3pt}
% Row and rule spacing are tightened so that the manuscript stays at 8 pages with this font size
% (with the default spacing it runs to 9 pages; \arraystretch 0.88 is already too loose).
\renewcommand{\arraystretch}{0.85}\aboverulesep=0.2mm \belowrulesep=0.2mm
\begin{tabular}{@{}clcccc@{}}
\toprule
\rule[-3.9pt]{0pt}{11.1pt}$N$ & Condition & $\rho$ & Avg.\ Exp.\ Rate & Success Rate (\%) & Overlap Ratio \\  % the strut pads the header row; a 13.6pt strut already gives 9 pages
\midrule
\multirow{7}{*}{$2$}
 & Nominal       & 0.086 & 0.962 & 67.0 & 0.561 $\pm$ 0.297 \\
 & Drop $p=.50$  & 0.033 & 0.968 & 62.0 & 0.544 $\pm$ 0.175 \\
 & Drop $p=.75$  & 0.017 & 0.964 & 54.0 & 0.554 $\pm$ 0.133 \\
 & Drop $p=1$    & 0    & 0.947 & 43.0 & 0.683 $\pm$ 0.122 \\
 & Range $r=7.5$\,m & 0.089 & 0.961 & 62.0 & 0.573 $\pm$ 0.297 \\
 & Range $r=5$\,m   & 0.085 & 0.961 & 58.0 & 0.594 $\pm$ 0.292 \\
 & Range $r=2.5$\,m & 0.031 & 0.949 & 44.0 & 0.660 $\pm$ 0.187 \\
\midrule
\multirow{7}{*}{$4$}
 & Nominal       & 0.198 & 0.996 & 92.0 & 0.689 $\pm$ 0.161 \\
 & Drop $p=.50$  & 0.091 & 0.995 & 92.0 & 0.699 $\pm$ 0.131 \\
 & Drop $p=.75$  & 0.055 & 0.991 & 92.0 & 0.713 $\pm$ 0.105 \\
 & Drop $p=1$    & 0    & 0.973 & 68.0 & 0.810 $\pm$ 0.094 \\
 & Range $r=7.5$\,m & 0.198 & 0.985 & 85.0 & 0.736 $\pm$ 0.168 \\
 & Range $r=5$\,m   & 0.202 & 0.989 & 88.0 & 0.719 $\pm$ 0.176 \\
 & Range $r=2.5$\,m & 0.094 & 0.981 & 76.0 & 0.794 $\pm$ 0.154 \\
\midrule
\multirow{7}{*}{$8$}
 & Nominal       & 0.313 & 0.999 & 98.0 & 0.845 $\pm$ 0.110 \\
 & Drop $p=.50$  & 0.183 & 0.999 & 98.0 & 0.827 $\pm$ 0.123 \\
 & Drop $p=.75$  & 0.102 & 0.997 & 97.0 & 0.808 $\pm$ 0.069 \\
 & Drop $p=1$    & 0    & 0.986 & 83.0 & 0.835 $\pm$ 0.074 \\
 & Range $r=7.5$\,m & 0.313 & 0.999 & 99.0 & 0.839 $\pm$ 0.134 \\
 & Range $r=5$\,m   & 0.324 & 0.999 & 98.0 & 0.855 $\pm$ 0.117 \\
 & Range $r=2.5$\,m & 0.153 & 0.993 & 88.0 & 0.839 $\pm$ 0.116 \\
\bottomrule
\end{tabular}}
\vspace{-15pt}
\end{table}

\subsection{Quantitative Comparison}
\begin{figure}[t]
\captionsetup{skip=2pt}
\centering
\includegraphics[width=\columnwidth]{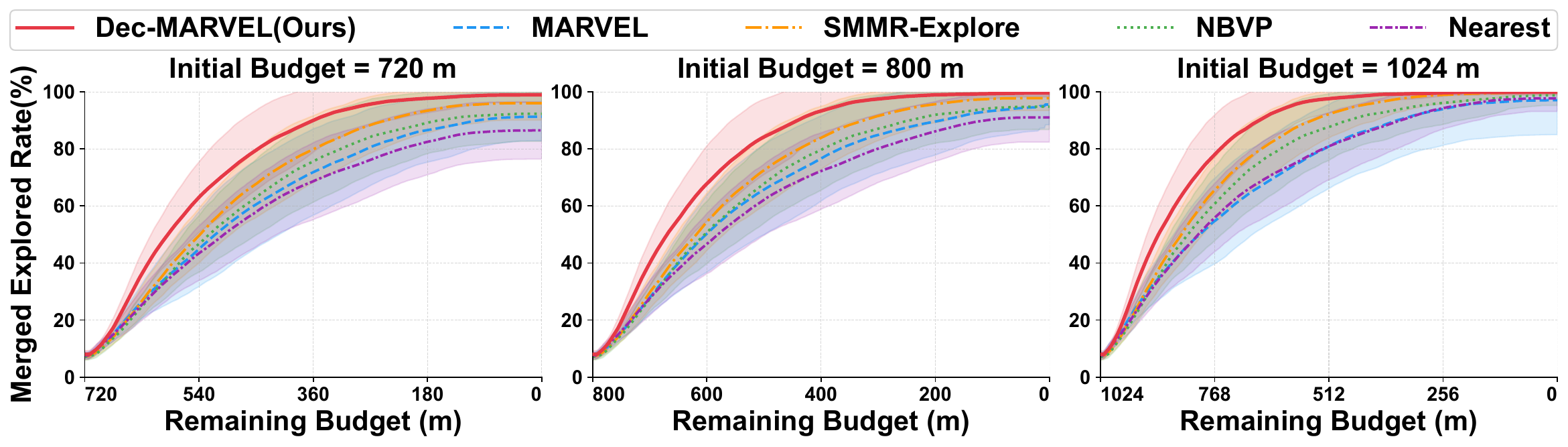}
\caption{Merged exploration rate for $N=4$ agents under budgets $B \in \{720,800,1024\}$\,m. \revision{Dec-MARVEL overtakes all baselines within the first third of the budget and maintains the highest merged coverage thereafter, indicating improved coverage efficiency before mission completion. SMMR-Explore and MARVEL denote the adapted variants of Sec.~\ref{sec:baselines}.}}
\label{fig:rate_history_4agent}
\vspace{-20pt}
\end{figure}

\vspace{-3pt}
\revision{Across the nine budget--team settings in Table~\ref{tab:main_results}, Dec-MARVEL achieves the highest or tied-highest exploration and success rates and the shortest reported distance to 99\% merged coverage among successful trials. Relative to the strongest baseline in each setting, exploration rate improves by 2.1\% on average and completion distance falls by 14.6\%. The success gains are pronounced at 720\,m: 29, 27, and 9 percentage points for 2, 4, and 8 robots, respectively; at 800\,m, the corresponding gains are 36, 24, and 5 points. As coverage approaches saturation at 1024\,m, Dec-MARVEL ties the best success rate for 2 and 8 robots, yet reaches 99\% coverage with 11.5\% less travel than the strongest baseline for 8 robots. Dec-MARVEL has the lowest FoV overlap in five of the six 2- and 4-robot settings; SMMR-Explore (adapted) and NBVP achieve lower overlap with 8 robots. Fig.~\ref{fig:rate_history_4agent} shows faster early and middle coverage for four robots, while the accompanying video illustrates complementary sensing footprints in a 10-robot 3D simulation~\cite{cao2022autonomous}.}

\subsection{Ablation and Robustness Studies}
\label{sec:ablation_robustness}
\vspace{-3pt}
Table~\ref{tab:ablation} evaluates the main training components at $B=800$\,m with $N=4$ robots. Removing the TOP critic~\eqref{top_critic} increases the distance to $99\%$ merged coverage \revision{from $403.56$\,m to $501.83$\,m ($+24.4\%$)}. Removing the mixture-based budget curriculum \eqref{eq:budget_curriculum}--\eqref{eq:mixture_sampling} and phase-conditioned critic~\eqref{eq:phase_conditioned_critic} increases completion distance \revision{by $7.3\%$ and $4.3\%$}, respectively.

% Table~\ref{tab:breakdown_budget} evaluates robustness to robot failures without retraining by randomly disabling one or two robots from a four-robot team between decision steps $30$ and $128$, after which failed robots stop moving and sensing. Dec-MARVEL degrades gracefully: with two failed robots, the average exploration rate remains $0.954$, $0.966$, and $0.969$ for budgets $720$, $800$, and $1024$\,m, respectively, while success remains $69$--$82\%$ despite fewer active robots completing the same coverage objective. This indicates that Dec-MARVEL does not rely on fixed explicit assignments and can adapt its local decisions when teammates disappear from the mission.

\revision{Table~\ref{tab:breakdown_budget} tests robot failures without retraining. With two of four robots disabled, Dec-MARVEL retains 0.961--0.985 average coverage and 78--91\% success across the tested budgets. This tests loss of active robots, not missed observations of robots that continue exploring.}
% The performance loss is proportional to the reduction in sensing capacity rather than a breakdown in coordination, as robots naturally redistribute exploration from local observations and passively observed teammate trajectories, without replanning or role reassignment. These results demonstrate robustness to partial team failures, supporting long-duration decentralized missions with uncertain robot availability.

\revision{Table~\ref{tab:visibility_robustness} evaluates the sensitivity to sparse teammate observations with the map detection unchanged. In the nominal setting, robots observe at least one teammate during 8.6\%, 19.8\%, and 31.3\% of exploration-mode decision steps for $N=2,4,8$, respectively. With 50\% detection dropout, success differs from nominal by at most 5 percentage points. Under complete dropout ($p=1$), success falls by 24, 24, and 15 points, respectively, while average exploration remains above 0.94. Complete observation dropout also increases overlap by about 0.12 for $N=2$ and $N=4$, while overlap changes little for $N=8$, suggesting that teammate observations are particularly useful for reducing redundant sensing in smaller teams. Restricting teammate detection to 7.5 or 5\,m keeps success within 9 percentage points of nominal, while reducing the range to 2.5\,m decreases success by 10--23 points. This suggests that the spatial distribution of observations, in addition to their frequency, affects coordination. In a separate perturbation test without retraining, Gaussian range/bearing/heading noise up to $0.229r$, $10^\circ$, and $20^\circ$ standard deviation changes success by at most 4 percentage points, while nearest-node associations change for at most 9.5\% of observations with four robots.}

\subsection{Real-World Experiments}
\label{sec:real-world_experiments}
\begin{figure}[t]
\captionsetup{skip=2pt}
\centering
\includegraphics[width=0.95\columnwidth]{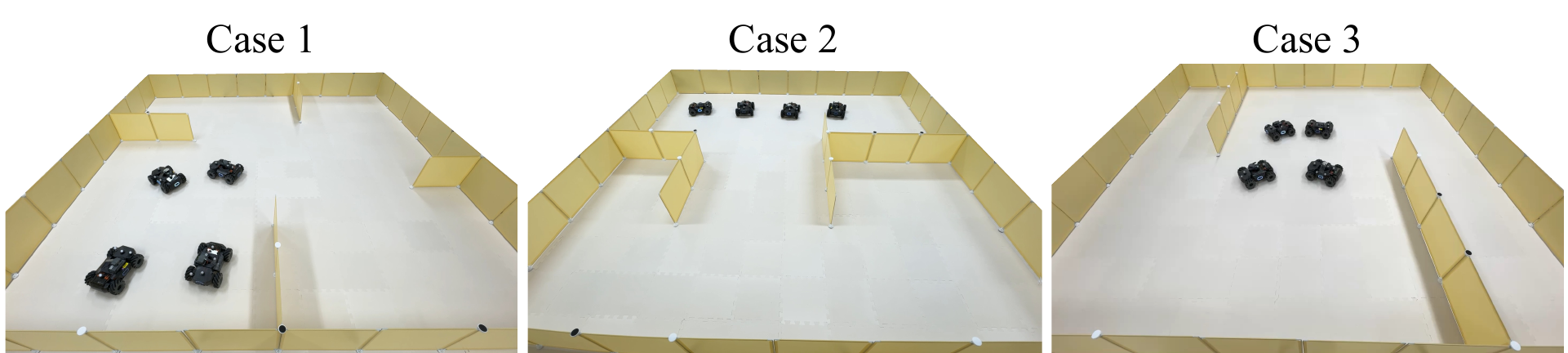}
\caption{Real-world experiments in three arenas. Four robots replay Dec-MARVEL waypoint-heading plans using motion-capture feedback.}
\label{fig:real_experiments}
\vspace{-20pt}
\end{figure}

\vspace{-3pt}
\revision{We replay plans generated in simulation, without retraining, on four DJI RoboMaster EP robots in three $3.6$\,m\,$\times$\,$3.6$\,m arenas (Fig.~\ref{fig:real_experiments}) to test physical execution within a travel budget. Given the arena size and robot footprint, we reduce the sensing range to $0.67$\,m ($120^\circ$ FoV) and the budget to $9.33$\,m, below the training range. Local maps are rendered from the floor plan at each simulated pose, and teammate position, heading, and identity are read noise-free from the simulator only under the same visibility conditions; no onboard perception is used. The policy instances share a workstation (AMD Ryzen Threadripper 7960X CPU, NVIDIA GeForce RTX 3090 GPU), but each receives only its own map, budget, and visible teammate trajectories; planning averages $0.3$\,s per cycle per robot.}

% \revision{Unlike in simulation, robot footprints are non-negligible in these arenas, and robots can deadlock in narrow passages. The hardware experiment therefore departs from the communication-free setting in two ways. First, plans are generated with the centralized arbiter of Sec.~\ref{sec:soft_actor_critic} instead of the rule in Sec.~\ref{sec:runtime_feasibility}. Second, during execution, a wait-and-resume rule on the workstation uses the robots' motion-capture positions and next waypoints, letting the robot closest to its next waypoint move first while the others wait and then resume, and an artificial potential field (APF) keeps robots apart. Both are applied identically to every method and provide no additional inputs to the policies; the execution layer changes only execution timing and local motion, not the planned waypoint-heading sequences. All robots completed their plans without collision and within budget, and in simulations of the same arenas, Dec-MARVEL achieved higher exploration rates than the baselines. Motion-capture positions are used only for trajectory tracking and by this execution layer, and are never used to generate the teammate observations provided to the policy.}

\revision{Because robot footprints are non-negligible in the small hardware arenas, plan generation uses the centralized training arbiter and execution uses a common workstation-based wait-and-resume rule with APF collision avoidance. Ground-truth robot positions from motion capture, including teammate positions, are available only to this execution layer for trajectory tracking and collision avoidance; they are not provided to the policy or used to generate teammate observations. These mechanisms are applied identically to all methods and modify only execution timing and local motion, not policy inputs or planned waypoint-heading sequences. All robots completed their plans without collision and within budget.}

% \revision{To test physical execution within a travel budget, we replay plans generated in simulation, without retraining, on four DJI RoboMaster EP robots in three $3.6$\,m$\times3.6$\,m arenas (Fig.~\ref{fig:real_experiments}). The simulator renders each robot's local map from the floor plan at its simulated pose, using a $120^\circ$ FoV, $0.67$\,m range, and line-of-sight checks. Teammate position, heading, and identity come from noise-free simulated state only when the teammate's center satisfies the same visibility conditions; no onboard perception is used. The policy instances share a workstation process (AMD Ryzen Threadripper 7960X CPU, NVIDIA GeForce RTX 3090 GPU), but each receives only its own map, budget, and visible teammate trajectories. Planning averages $0.3$\,s per cycle per robot.}

% \revision{The $9.33$\,m arena budget is below the training range, and robot footprints are non-negligible. Motion capture is used only for low-level trajectory tracking, not as policy input. A common artificial potential field and wait-and-resume rule is applied to Dec-MARVEL and all baselines for collision avoidance. Robot positions are provided by the simulator for convenience, but we note that this could be estimated using onboard vision. Since this safety layer is shared across all methods and does not affect planning decisions, it does not bias the comparison. All robots completed their plans without collision and within budget, while Dec-MARVEL also achieved higher exploration rates in simulation.}

\section{Conclusion}
\revision{Dec-MARVEL shows that a decentralized, communi\-cation-free policy can combine local frontiers, remaining budget, and observed teammate trajectories to make coordinated, budget-aware waypoint-heading decisions. Across the tested budgets and team sizes, it achieves the highest or tied-highest exploration and success rates with the shortest reported travel to 99\% coverage among successful trials, and the lowest or near-lowest sensing overlap for teams of two and four robots. Because each robot considers only the teammates it observes, two robots can still select the same waypoint; physical collision avoidance is therefore delegated to a low-level controller. Future work will study onboard perception with real detection and identity errors, identity-based priority for contested waypoints, and integration with decentralized collision avoidance.}

\bibliographystyle{IEEEtran}
\bibliography{references}

\end{document}